\documentclass[conference]{IEEEtran}
\IEEEoverridecommandlockouts
\usepackage[table]{xcolor}

\usepackage{cite}
\usepackage{amsmath,amssymb,amsfonts, mathtools}
\usepackage{algorithmic}
\usepackage{graphicx}
\usepackage{textcomp}
\usepackage{xcolor}
\def\BibTeX{{\rm B\kern-.05em{\sc i\kern-.025em b}\kern-.08em
    T\kern-.1667em\lower.7ex\hbox{E}\kern-.125emX}}
    
\newcommand{\comment}[1]{}

\usepackage{bm} 
\usepackage{fixmath} 
\newcommand*{\mbf}{\boldsymbol}

\usepackage{bbm} 
\usepackage{comment}
\usepackage{tabularx}
\usepackage{booktabs}
\definecolor{lightgray}{gray}{0.97}
\usepackage{subfig}
\usepackage{multirow}
\usepackage{hhline}

\begin{document}

\title{EventScore: An Automated Real-time Early Warning Score for Clinical Events 
}

\author{\IEEEauthorblockN{Ibrahim Hammoud\IEEEauthorrefmark{1},
Prateek Prasanna\IEEEauthorrefmark{2}, 
  IV Ramakrishnan\IEEEauthorrefmark{3}, 
 Adam Singer\IEEEauthorrefmark{4},
  Mark Henry\IEEEauthorrefmark{5}
  and Henry Thode\IEEEauthorrefmark{6} }
\IEEEauthorblockA{
\IEEEauthorrefmark{1}\IEEEauthorrefmark{3}Department of Computer Science\\
\IEEEauthorrefmark{2}Department of Biomedical Informatics\\
\IEEEauthorrefmark{4}\IEEEauthorrefmark{5}\IEEEauthorrefmark{6}Department of Emergency Medicine\\
Stony Brook, NY, USA \\
Email: \{\IEEEauthorrefmark{1}ihammoud, \IEEEauthorrefmark{3}ram\}@cs.stonybrook.edu,  \IEEEauthorrefmark{2}prateek.prasanna@stonybrook.edu \\ \{\IEEEauthorrefmark{4}adam.singer, \IEEEauthorrefmark{5}mark.henry, \IEEEauthorrefmark{6}henry.thode\}@stonybrookmedicine.edu}}

\maketitle

\begin{abstract}
Early prediction of patients at risk of clinical deterioration can help physicians intervene and alter their clinical course towards better outcomes. In addition to the accuracy requirement, early warning systems must make the predictions early enough to give physicians enough time to intervene. Interpretability is also one of the challenges when building such systems since being able to justify the reasoning behind model decisions is desirable in clinical practice. In this work, we built an interpretable model for the early prediction of various adverse clinical events indicative of clinical deterioration. The model is evaluated on two datasets and four clinical events. The first dataset is collected in a predominantly COVID-19 positive population at Stony Brook Hospital. The second dataset is the MIMIC III dataset. The model was trained to provide early warning scores for ventilation, ICU transfer, and mortality prediction tasks on the Stony Brook Hospital dataset and to predict mortality and the need for vasopressors on the MIMIC III dataset. Our model first separates each feature into multiple ranges and then uses logistic regression with lasso penalization to select the subset of ranges for each feature. The model training is completely automated and doesn't require expert knowledge like other early warning scores. We compare our model to the Modified Early Warning Score (MEWS) and quick SOFA (qSOFA), commonly used in hospitals. We show that our model outperforms these models in the area under the receiver operating characteristic curve (AUROC) while having a similar or better median detection time on all clinical events, even when using fewer features. Unlike MEWS and qSOFA, our model can be entirely automated without requiring any manually recorded features. We also show that discretization improves model performance by comparing our model to a baseline logistic regression model.

\end{abstract}

\section{Introduction}
Rule-based early warning score systems are developed to predict patients at risk of deterioration. Although developing such systems is not entirely automated and often requires healthcare professionals' input, one of their most significant advantages is that they are interpretable and verifiable. For instance, let's take an example of a system that raises an alarm if two or more of the patient's vital signs (such as heart rate and body temperature) are outside the normal range. Such a system is interpretable because we can understand the cause of an alarm every time the alarm is raised. Verifiability in such a system stems from the fact that for any combination of vital sign values, we can understand how the system will behave. Even when complex data-driven models can outperform rule-based early warning systems, physicians do not feel confident using them. This is because most data-driven models have parameters that are hard or too complex to interpret for the end user (deep-learning models, SVMs, random forests, gradient boosting trees). While on the other end, simpler linear models fail to capture non-linear patterns in a given feature.

Severity score systems can be separated into two categories. The first category includes intensive care unit (ICU) scoring systems. In this category, a final risk score is assigned to each patient after 24 hours of admission to the ICU. The worst value of each feature (according to the given severity score) during that time is then selected to assign a risk score. The second category includes early warning scores that are used in the Emergency Department (ED) as well in the ICU. Such systems continuously assign scores and are designed to raise alarms for medical care professionals for early intervention. Among the most commonly used scores in the first category are APACHE (I, II, III and IV) \cite{apache_1, apache_2, apache_3, apache_4}, SAPS (I, II, III) \cite{saps_1, saps_2, saps_3}, LODS \cite{lods}, OASIS \cite{oasis} and AutoScore \cite{autoscore}. APACHE was developed to predict hospital mortality. SAPS was then derived to simplify the APACHE score. While APACHE I, APACHE II, and SAPS were designed using a panel of experts, APACHE III and SAPS II were later developed using multiple logistic regression after separating each feature into multiple ranges based on expert advice. A dataset of 17,440 adult intensive care unit (ICU) admissions at 40 US hospitals was used in APACHE III. SAPS II uses locally weighted least squares (LOWESS) to create a univariate smoothed function of hospital mortality against a given variable. The ranges were then assigned via inspection. LODS uses the same dataset and method as SAPS II, however, for predicting organ dysfunction instead of mortality. SAPS III was later developed using the same modeling techniques as SAPS II, however, calibrated to new datasets that include ICUs from more countries and to adjust for the change in prevalence of major diseases and the changes in therapeutics. APACHE was later updated to version IV, and it expanded to more features and used restricted cubic regression splines combined with multiple logistic regression. Each variable is separated into multiple variables after choosing the cut points between the splines. The OASIS score uses data from 86 ICUs at 49 hospitals to predict mortality. A genetic algorithm is first used to select the subset of predictor variables to use. A particle swarm algorithm is then used to assign scores after separating each variable into deciles. AutoScore is an automated algorithm that uses logistic regression to generate a warning score given a dataset. The system first ranks features using a random forest. Features are also separated into ranges based on predefined percentiles. After that, each feature's mean in the last 24 hours is computed for each patient. The vector representing the mean value for every feature for a given patient is then used to predict the risk score. AutoScore was tested for mortality prediction using the MIMIC \cite{MIMIC_3} dataset. One advantage of the OASIS and AutoScore over the other scores is that they do not need expert knowledge to separate the features or decide on the cut points.

Although proven accurate in predicting clinical events, the above rule-based systems are designed as single point prediction systems. They are not designed as real-time early warning systems. 

Moving to the second category of severity scores, the Modified Early Warning Score (MEWS) was developed to detect deteriorating patients in a hospital ward \cite{mews}. It uses five physiological features that are repeatedly measured in the hospital. Using MEWS is associated with earlier admissions to the ICU. On the other hand, the Sepsis-related Organ Failure Assessment (SOFA) \cite{sofa} score is commonly used for the assessment of organ dysfunction. A higher SOFA score is associated with increased mortality. However, it requires lab values, which makes the system not suitable for repeated computation since lab measurements are not taken as frequently as vital signs. For this, the third International consensus definitions for sepsis and septic shock recommends using quick SOFA (qSOFA) \cite{sepsis3}, a modified simpler version of the SOFA score, to assess patients repeatedly. Such a score serves as an early warning that prompts physicians to investigate patients further.  

In this work, we focus on the second category of severity scores. We build a tool for generating an early warning score given a dataset and a clinical event without the need for expert knowledge. This is helpful since most of the systems we discussed that depend on expert knowledge take years to be updated and are carefully designed for specific clinical events. The rise of the COVID-19 pandemic increased the need for early warning scores that are tailored for new populations such as the COVID-19 patients. This, for instance, would help medical care professionals attend to the patients at risk early on while at the same time helping them estimate the amount of ICU beds and ventilators needed. 

Although other automated early warning scores exist, many of them use algorithms that make it harder for medical care professionals to interpret the model decisions. Moreover, many of those systems that use complex algorithms require more data, which is not desirable in a rising pandemic like COVID-19. State-of-the-art models are mainly tailored for specific clinical events, and it is unclear how well they generalize to other clinical events. On the other hand, our system produces an early warning score in a format that is familiar to medical care professionals and works well on different clinical events, even when the dataset is limited.

\section{Methods}

\subsection{Cohort}
The model was trained and tested on two datasets and different clinical events: Stony Brook Hospital dataset (ventilation, ICU transfer, and mortality) and MIMIC III dataset (mortality, vasopressor administration). For a given dataset and a clinical event, we define positive and negative patients to represent patients who eventually had/did not have the given clinical event, respectively. The number of positive and negative patients for each clinical event and dataset are displayed in Table \ref{tab:dataset}. 

The Stony Brook hospital data was collected retrospectively from February 7th, 2020, to May 4th, 2020. It includes patients throughout their hospital stay. Moreover, due to the COVID-19 pandemic, the Stony Brook hospital patient population was predominantly COVID-19 positive. For instance, the ventilation dataset contains 1,685 hospital admissions, out of which 203 resulted in the need for ventilation. Among the ventilated patients, 184 were COVID-19 positive. For the remaining 1,481 admissions that didn't have a ventilation event, 898 patients tested positive for COVID-19, 524 tested negative, and 59 had no tests performed. Since the Stony Brook hospital dataset contains patients whose discharge date is censored, we removed all negative patients who were not yet discharged at the end of the time frame used to collect the data. The ventilation event includes patients who were intubated after admission to the hospital and excludes patients who were intubated by the Emergency Department (ED) or by the Emergency Medical Services (EMS). The mortality event includes all patients who died at the hospital, excluding patients who died in the ED. The ICU transfer event contains patients who were transferred to the ICU at some point during their hospital stay. This excludes patients who were directly admitted to the ICU. 

Medical Information Mart for Intensive Care (MIMIC) III is a publicly available database of deidentified health-related data associated with over 40,000 patients and over 60,000 ICU stays. The data were collected during those patients' stay in critical care units of the Beth Israel Deaconess Medical Center between 2001 and 2012. In this study, we only use the most recent data collected between 2008 and 2012, which has a different data management system than the previous data collected between 2001 and 2008. We didn't remove right-censored patients from the MIMIC III dataset since all patients were eventually discharged. For patients who had multiple ICU stays, we treated each ICU stay as a separate data point. 

For all datasets, we only included adult patients (age 18 or older). The Stony Brook hospital dataset contains 15 different physiological and lab features. On the other hand, for the MIMIC III dataset, we extracted eight physiological features. Table \ref{tab:features} shows the different features used in each dataset in addition to the average number of records of each feature per patients who have at least one recording of those features.

\begin{table}[h]
\def\arraystretch{1.5}
\resizebox{\columnwidth}{!}{
\begin{tabular}{|c|c|cc|cc|}
\cline{3-6}
 \multicolumn{1}{c}{}& \multicolumn{1}{c|}{} & \multicolumn{2}{c|}{Positive admissions} & \multicolumn{2}{c|}{All admissions} \\ \specialrule{.1em}{.0em}{.0em}
Dataset &  Clinical event& Total & \shortstack{COVID+} & Total & \shortstack{COVID+} \\ \specialrule{.1em}{.0em}{.0em}
\multirow{3}{1.4cm}{\shortstack{Stony Brook \\Hospital}} & Ventilation & 203& 184& 1,685 &1, 081\\ \cline{2-6}
& Transfer to ICU & 223 & 194 & 1,724 & 1,103\\ \cline{2-6}
& Mortality & 168 & 137 &1,649 & 1,034 \\ \specialrule{.1em}{.0em}{.0em}
\multirow{2}{1.4cm}{MIMIC III} & Mortality & 2,267& -  &23,362 & -\\ \cline{2-6}
& \shortstack{Vasopressor \\ administration} & 6360 & - &23,362 & -\\ \specialrule{.1em}{.0em}{.0em}
\end{tabular}
}
\caption{Number of hospital admissions for each dataset and clinical event. COVID+ indicates the number of hospital admissions associated with COVID-19 positive patients. MIMIC III dataset was extracted between 2008 and 2012 and thus it contains no COVID-19 patients.}
\label{tab:dataset}
\end{table}

~

\begin{table*}[!h]
\center
 \rowcolors{3}{}{lightgray}
\begin{tabular}{|cc|cc|cc|}
\cline{3-6} 
\multicolumn{2}{c|}{}& \multicolumn{2}{c|}{Stony Brook Hospital Dataset} & \multicolumn{2}{c|}{MIMIC III Dataset}  \\    \specialrule{.1em}{0em}{.1em}
        \shortstack{Feature} & \shortstack{Shorthand notation}  & \shortstack{Fraction available \\ per patient} & \shortstack{Average number of  \\ values per patient}  & \shortstack{Fraction available \\ per patient} & \shortstack{Average number of  \\ values per patient} \\
    \specialrule{.1em}{.1em}{.1em}
        
 Alanine aminotransferase &  ALT  & 0.98 & 5.7  & - & -\\
 Alert, Voice, Pain, Unresponsive scale&  AVPU  & 1.00 & 61.4 & - & - \\
C creative protein&  CRP  & 0.91 & 6.7  & - & -\\
Diastolic Blood Pressure & DBP &- & - & 1.00 & 82.95 \\
Glascow Coma Score & GCS & -& -& 1.00 & 24.13\\
Glucose & GLUCOSE & - & - & 0.99 & 18.49 \\
Ferritin&  FERR  & 0.82 & 6.0  & - & -\\
Fibrin D-dimer DDU&  DDIM  & 0.84 & 6.6  & - & -\\
Heart Rate &  HR  & 1.00 & 104.7  & 1.00 & 86.54\\
Lactate dehydrogenase&  LDH  & 0.89 & 5.5  & - & -\\
Leukocytes &  LEUKO  & 0.73 & 9.5  & - & -\\
Lymphocytes &  LYMPH  & 1.00 & 6.1  & - & -\\
Natriuretic peptide.B prohormone N-Terminal &  BNP  & 0.58 & 1.2  & - & -\\
Mean Blood Pressure & MBP & - & - & 1.00 & 84.68 \\
Oxygen Saturation in arterial blood &  O2SAT  & 1.00 & 107.9 & - & - \\
Oxygen Saturation & SPO2 & - & - & 1.00 & 85.01 \\
Procalcitonin&  PROCAL  & 0.91 & 5.4  & - & -\\
Respiratory rate &  RR  & 1.00 & 96.6  & 1.00 & 86.48\\
Systolic Blood Pressure &  SBP  & 1.00 & 78.4  & 1.00 & 82.97\\
Temperature &  TEMP  & 1.00 & 37.7  & 0.99 & 24.60\\
Troponin T.cardiac &  TNT  & 0.85 & 2.5 & - & - \\
         \bottomrule
    \end{tabular}
    \caption{Features used in each dataset. Average number of records per patient was only calculated for patients who had at least one recording of the feature.}
    \label{tab:features}
\end{table*}

\begin{figure}[h]
\centering
\includegraphics[width=1.0\linewidth]{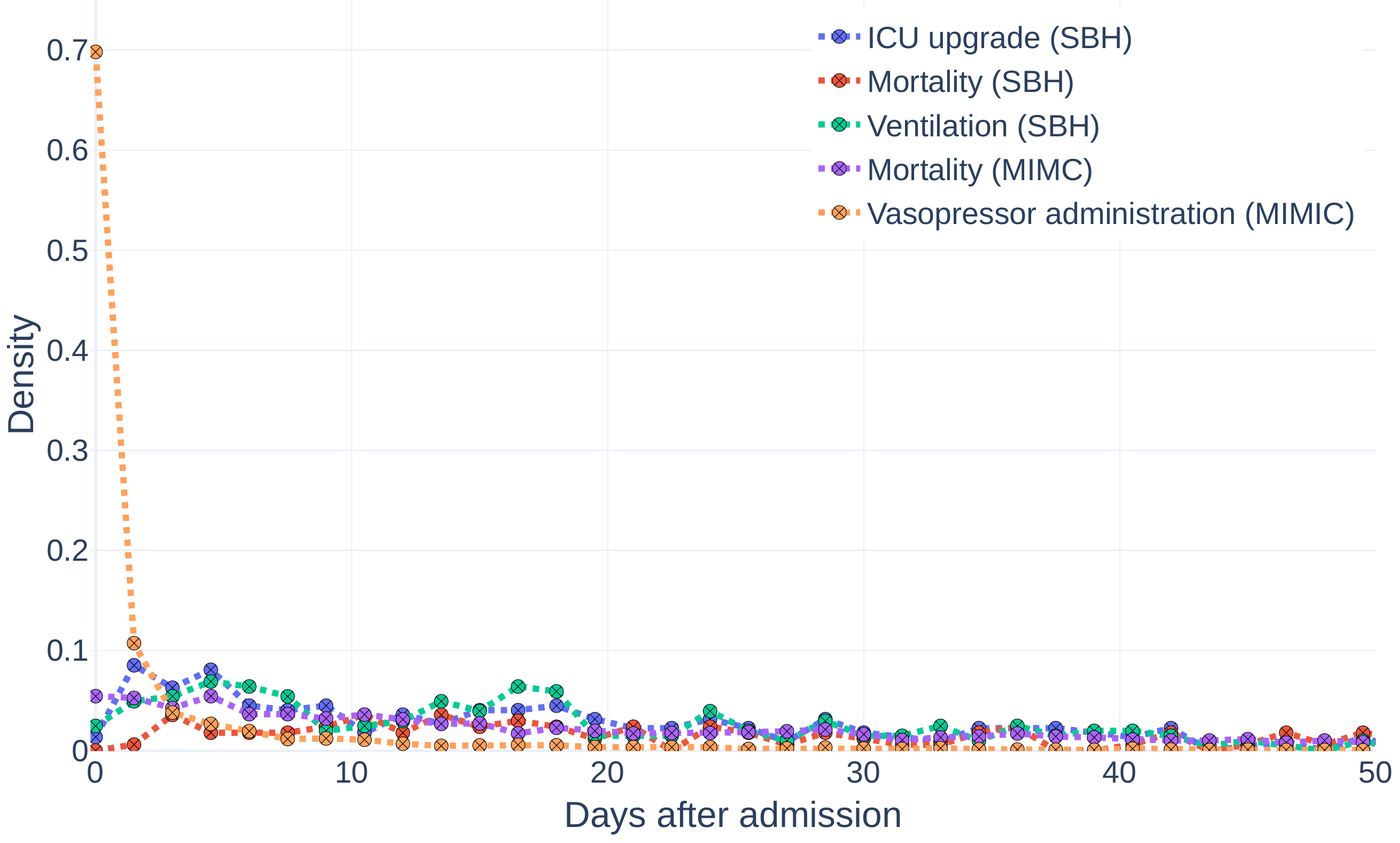}
\caption{ The distribution of the patient clinical event times in days.}
\label{fig:density}
\end{figure}

Figure \ref{fig:density} shows the density for each clinical event in different datasets. While the distributions are long tailed in general, we can notice that most vasopressor administrations were within the first 6 hours after the start of the ICU stay.

\subsection{Model}
The data of each hospital stay $i$ is represented as a series of vectors $\mbf{x_{ij}}, ... \mbf{x_{in_i}}$, where each vector $\mbf{x_{ij}}$ represents the patient's state at timestamp $j$. We bin the records of each hospital admission $i$ into half-hour windows, and we denote by $n_i$ the number of resulting timestamps. Each vector represents the most recent patient records within that half-hour window. Each feature's missing values are imputed with the last available value. Moreover, for each patient, we initially impute each feature's value with the median value, where the median value is computed after taking the median value from each patient for that feature. Taking the median at the patient level ensures that each patient can only contribute one value to this median, thus preventing long hospital admissions from dominating the final value used for imputation. Using the median instead of the mean helps against cases where some features can have large deviations in some rare cases, which can impact the mean but not the median. 

For each hospital stay $i$, we denote by $y_i$ the binary label indicating whether the patient eventually developed the positive clinical event within that stay. Since signs of deterioration increase as we approach the time of the event among positive patients, we rescale the weights of each data-point in their hospital stay to increase linearly between 0 and 1 over the last 72 hour period before the positive event onset. Data-points for those hospital admissions that happen before 72 hours of the positive event time were excluded from the training set but included in the test set. Hospital admissions (positive and negative) have different lengths. To avoid long hospital admissions from dominating the loss function, and to give equal contribution for each hospital stay, we re-weight each patient's vectors in that hospital stay such that they sum up to one.

 We use logistic regression with lasso regularization for the given prediction task:
 \begin{align*}
 \begin{multlined}[\columnwidth]
\min_{(\beta_0, \mbf{\beta}) \in \mathbb{R}^{p+1}} \lambda ||\mbf{\beta}||_1 - \\ \left[\frac{1}{N} \sum_{i=1}^N \sum_{j=1}^{n_i}w_{ij}\Big( y_i \cdot h(\mbf{x_{ij}}) - \log (1+e^{h(\mbf{x_{ij})}})\Big)\right]
\end{multlined}
\end{align*}

Where $h(x) = \beta_0 + \mbf{x}^T \mbf{\beta}$ and $\sum_{j=1}^{n_i} w_{ij} = 1$
\\ \\
 Lasso regularization allows the model to zero out coefficients that do not contribute to the prediction task. This helps in improving interpretability by making the model solution simpler. We use 5-fold cross-validation on the training set to select the hyper-parameter $\lambda$ controlling the regularization strength. After that, we report our model performance on the final testing set.

For baseline logistic regression, each vector $\mbf{x_{ij}}$ contains the original numeric features contained in the dataset. For EventScore, on the other hand, each feature was discretized into multiple ranges. Thus $\mbf{x_{ij}}$ is a multi-hot encoding representing the feature ranges of the measurements the patient had at time step $j$. This is done for each feature separately by selecting splits that maximize the information gain using the Classification And Regression Tree (CART) \cite{cart} algorithm on the training set. The information gain is weighted using the same weighting scheme for positive and negative patients as before; however, at a univariate level. Patients who never had records for the given feature were excluded when identifying the ranges.

\begin{table*}
\Huge
\def\arraystretch{1.6}
\rowcolors{6}{lightgray}{}
\resizebox{\textwidth}{!}{
\begin{tabular}{!{\vrule width 0.15em}c!{\vrule width 0.15em}c!{\vrule width 0.05em}c!{\vrule width 0.05em}c!{\vrule width 0.05em}c!{\vrule width 0.15em}c!{\vrule width 0.05em}c!{\vrule width 0.05em}c!{\vrule width 0.05em}c!{\vrule width 0.15em}c!{\vrule width 0.05em}c!{\vrule width 0.05em}c!{\vrule width 0.05em}c!{\vrule width 0.15em}c!{\vrule width 0.05em}c!{\vrule width 0.05em}c!{\vrule width 0.05em}c!{\vrule width 0.15em}c!{\vrule width 0.05em}c!{\vrule width 0.05em}c!{\vrule width 0.05em}c!{\vrule width 0.15em}}
\cline{2-21}\multicolumn{1}{c!{\vrule width 0.15em}}{} & \multicolumn{20}{c!{\vrule width 0.15em}}{Dataset} \\  \hline
&\multicolumn{4}{c!{\vrule width 0.15em}}{Mortality (SBH)}&\multicolumn{4}{c!{\vrule width 0.15em}}{Ventilation (SBH)}&\multicolumn{4}{c!{\vrule width 0.15em}}{ICU transfer (SBH)}&\multicolumn{4}{c!{\vrule width 0.15em}}{Vasopressors (MIMIC)}&\multicolumn{4}{c!{\vrule width 0.15em}}{Mortality (MIMIC)}\\
\cline{2-21}
&AUC & $t_{\text{qSOFA}}$ & $t_{\text{MEWS}}$ & p-value&AUC & $t_{\text{qSOFA}}$ & $t_{\text{MEWS}}$ & p-value&AUC & $t_{\text{qSOFA}}$ & $t_{\text{MEWS}}$ & p-value&AUC & $t_{\text{qSOFA}}$ & $t_{\text{MEWS}}$ & p-value&AUC & $t_{\text{qSOFA}}$ & $t_{\text{MEWS}}$ & p-value\\ \specialrule{.15em}{.0em}{.0em} 
qSOFA& 0.744 & 202.3 & -  & 0.0& 0.574 & 58.4 & -  & 0.0& 0.595 & 46.5 & -  & 0.0& 0.299 & 9.5 & -  & 0.0& 0.706 & 66.4 & -  & 0.0\\
Logistic Regression [qSOFA*]& 0.852 & 195.2 & 189.8  & 0.015& 0.725 & 61.3 & 30.7  & 0.35& 0.715 & 44.0 & 29.3  & 0.029& 0.36 & 4.7 & 2.8  & 0.0& 0.791 & 62.9 & 72.3  & 0.0\\
Logistic Regression [qSOFA]& 0.872 & 202.3 & 197.3  & 0.114& 0.724 & 62.6 & 30.7  & 0.3& \textbf{0.718} & 44.0 & 29.3  & 0.014& 0.36 & 4.7 & 2.8  & 0.0& 0.812 & 65.2 & 72.1  & 0.01\\
EventScore [qSOFA*]& 0.874 & 202.3 & 117.0  & 0.019& \textbf{0.729} & 62.6 & 30.2  & -& 0.696 & 44.0 & 26.2  & -& 0.5 & - & 2.8  & 0.0& 0.793 & 62.6 & 72.1  & 0.0\\
EventScore [qSOFA]& \textbf{0.89} & 202.3 & 164.0  & -& \textbf{0.729} & 62.6 & 30.2  & -& 0.696 & 44.0 & 26.2  & -& \textbf{0.516} & - & 2.8  & -& \textbf{0.822} & 64.2 & 72.0  & -\\
\specialrule{.15em}{.0em}{.0em}  MEWS& 0.855 & - & 166.9  & 0.011& 0.675 & - & 37.0  & 0.001& 0.663 & - & 36.4  & 0.0& 0.302 & - & 5.7  & 0.0& 0.79 & - & 72.1  & 0.0\\
Logistic Regression [MEWS*]& 0.849 & 199.0 & 166.9  & 0.005& 0.73 & 61.3 & 31.4  & 0.08& 0.724 & 47.5 & 29.4  & 0.026& 0.36 & 4.7 & 2.8  & 0.0& 0.777 & 66.5 & 72.6  & 0.0\\
Logistic Regression [MEWS]& 0.874 & 200.8 & 180.5  & 0.057& 0.727 & 63.2 & 31.1  & 0.052& 0.728 & 47.5 & 30.6  & 0.052& 0.36 & 4.7 & 2.8  & 0.0& 0.822 & 68.4 & 72.1  & 0.0\\
EventScore [MEWS*]& 0.878 & 203.5 & 150.2  & 0.063& \textbf{0.753} & 63.8 & 45.0  & -& \textbf{0.75} & 52.0 & 32.8  & -& 0.568 & 2.7 & 2.9  & 0.001& 0.821 & 67.7 & 72.1  & 0.0\\
EventScore [MEWS]& \textbf{0.893} & 202.3 & 168.2  & -& \textbf{0.753} & 63.8 & 45.0  & -& \textbf{0.75} & 52.0 & 32.8  & -& \textbf{0.575} & 2.7 & 2.9  & -& \textbf{0.84} & 66.2 & 72.1  & -\\
\specialrule{.15em}{.0em}{.0em}  Logistic Regression [all features]& 0.93 & 203.7 & 165.6  & 0.116& 0.749 & 61.3 & 31.4  & 0.127& 0.753 & 52.0 & 36.0  & 0.211& 0.389 & 4.5 & 2.8  & 0.0& 0.866 & 64.6 & 70.6  & 0.002\\
EventScore [all features]& \textbf{0.944} & 211.9 & 179.5  & -& \textbf{0.769} & 65.6 & 41.2  & -& \textbf{0.768} & 52.0 & 37.7  & -& \textbf{0.635} & 2.6 & 2.9  & -& \textbf{0.881} & 62.9 & 72.1  & -\\
\specialrule{.15em}{.0em}{.0em} 
\end{tabular}}

\caption{Performance of the different models across different datasets and clinical events. [Severity score] indicates that a model was trained using only the features of the given severity score. [Severity score*] is similar, with an added restriction of further excluding features that need human input. $t_{\text{qSOFA}}$ and $t_{\text{MEWS}}$ represent the median detection time when measured at the specificity corresponding to a qSOFA threshold of 2 and a MEWS threshold of 5, respectively. $p$-values  were calculated based on the AUC of the model in a given row against the AUC of the corresponding EventScore model in its row group (one of EventScore[qSOFA], EventScore[MEWS] and EventScore[All features]).}
\label{tab:results_final}
\end{table*}

\section{Results}

We divide all datasets into 60\% training and 40\% testing. For each clinical event and dataset, we extract ranges using the training set. After that, we select the best L1 regularization strength $\lambda$ using five-fold cross-validation on the training set. 

Also, to ensure a fair comparison against MEWS and qSOFA, we run some experiments to train our model on only the given features of those early warning scores. We also run other experiments where we add more features to the model to illustrate the model performance improvement as we increase the number of features used. We define the median detection time as the median number of hours prior to the event when the first alarm is raised for each positive patient correctly predicted at a given threshold for a particular model. MEWS and qSOFA have integer scores (qSOFA between 0 and 3, MEWS between 0 and 14), making it difficult to fix the false-positive rate to compare both systems across different datasets. This is because a prediction threshold in one system will be mapped to a false-positive rate, which in turn might not correspond to an actual threshold in the other system. This is also true when trying to find a fixed false-positive rate within the same system across different datasets. The unique set of false-positive rates is a property of the dataset and the clinical event used. Thus, to compare the data-driven models to qSOFA and MEWS, we first extract the false-positive rates from each system at thresholds used in the literature to predict patients at risk of deterioration. Namely, we use a threshold of 5 for MEWS and 2 for qSOFA. We then compute the median detection times $t_{\text{MEWS}}$ and $t_{\text{qSOFA}}$ for EventScore and logistic regression at the false-positive rates of MEWS and qSOFA at the given thresholds within the given dataset and clinical event. 

Table \ref{tab:results_final} summarizes the results across different datasets and clinical events. All AUC values are computed at the hospital stay level. $p$-values were computed using DeLong's algorithm \cite{delong, delong_fast} by comparing the AUC of every model to the AUC of EventScore[qSOFA], EventScore[MEWS] or EventScore[All features] depending on the subset of features the models is using.

The results in the table show that in all clinical events, EventScore outperforms MEWS and qSOFA when using the same subset of features. All improvements in AUC over MEWS and qSOFA are statistically significant. Moreover, the results show that EventScore also outperforms these systems when using fewer features by excluding GCS (or AVPU) score, which requires medical staff to observe the patient and record values manually. This helps make the system completely automated. Simultaneously, we can see that EventScore's early prediction quality is equal to those metrics. This is evident when comparing the median detection time $t_{\text{MEWS}}$ of EventScore[MEWS] and MEWS and the median detection time $t_{\text{qSOFA}}$ of EventScore[qSOFA] and qSOFA. Like qSOFA, the average of the median times across all events (except vasopressor administration since most models performed worse than random) of EventScore is 93 hours when using the qSOFA features. On average, EventScore[MEWS] detects clinical events a median of 79 hours in advance compared to MEWS's 78 hours. This demonstrates that the improvements in AUC compared to qSOFA and MEWS were not at the expense of a reduction in early prediction. 

Comparing the results to baseline logistic regression, we observe that EventScore outperforms logistic regression in all experiments except in one case in the ICU transfer dataset when using only the qSOFA features. For instance, using all the features we extracted from the MIMIC dataset, EventScore achieves an AUC of 0.881 on the mortality event compared to the baseline Logistic Regression's 0.866 ($p~$= 0.002). Similarly, on the ventilation event, EventScore achieves improved performance over Logistic Regression in all clinical events in all subsets of features. However, because Stony Brook hospital's dataset is much smaller than the MIMIC dataset, statistical significance could not be established in all cases. In some cases, EventScore did not select the features that require human input (AVPU/GCS) during training. For this reason, we can see that in some cases (such as the Stony Brook hospital ventilation dataset), the results of EventScore[qSOFA]/EventScore[MEWS] and EventScore[qSOFA*]/EventScore[MEWS*] are the same.

\begin{figure*}[!h]
\centering
\includegraphics[width=0.8\linewidth]{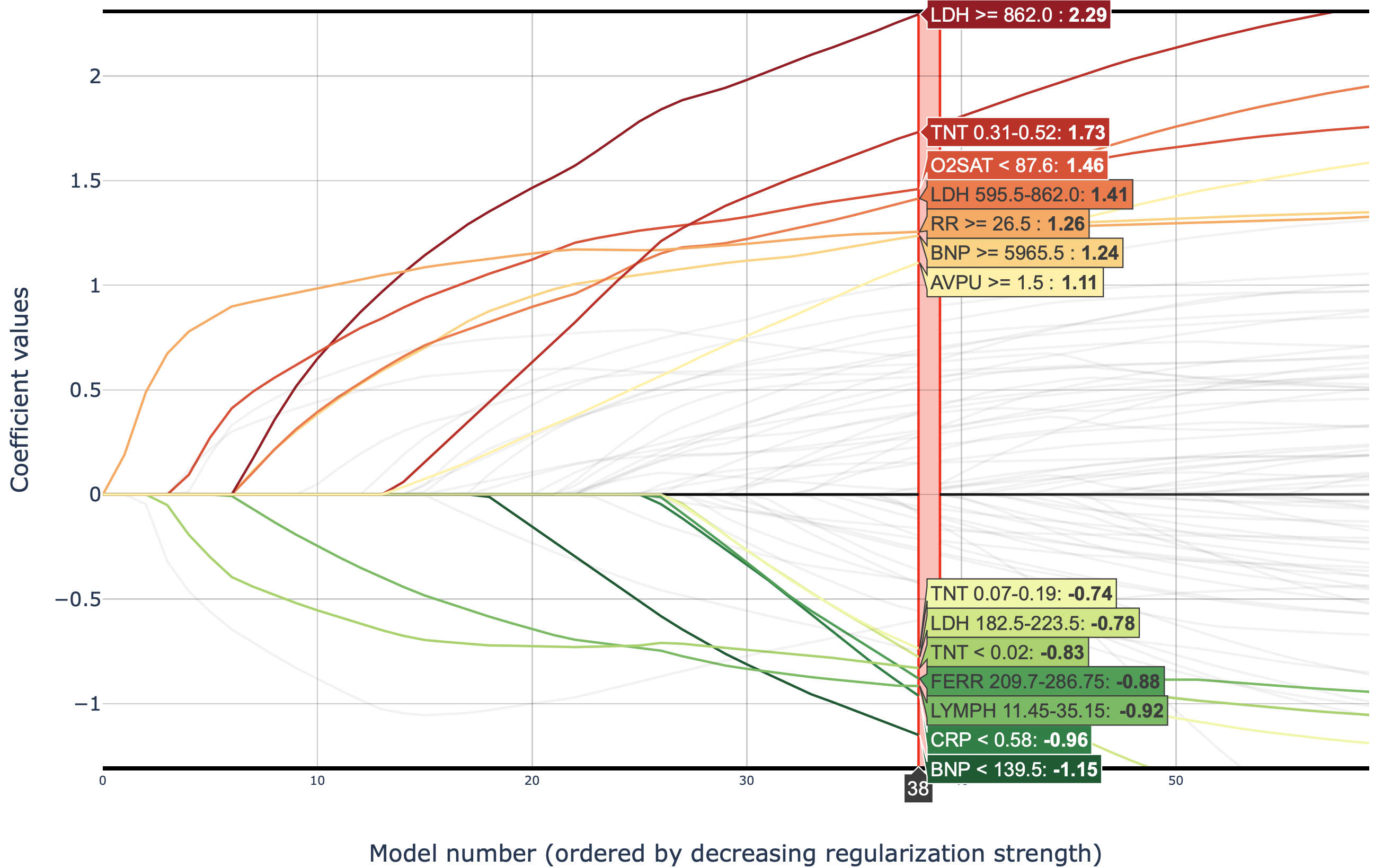}
\caption{ The graph shows how the model coefficients change as we increase the model complexity (from left to right) on the mortality event using all the 15 features from the Stony Brook Hospital dataset. Simpler models generalize better and have fewer coefficients, however, at the cost of reduced performance. More complex models use more features and perform better on the training set, however, at the cost of reduced performance. The red vertical line shows the model that was selected using five-fold cross-validation. Features with the highest coefficients (in absolute value) for the selected model are highlighted in red (increases the risk score) and green (decreases the risk score).  }
\label{fig:lasso_path}
\end{figure*}

\begin{figure}[!htbp]
\centering
\includegraphics[width=\linewidth]{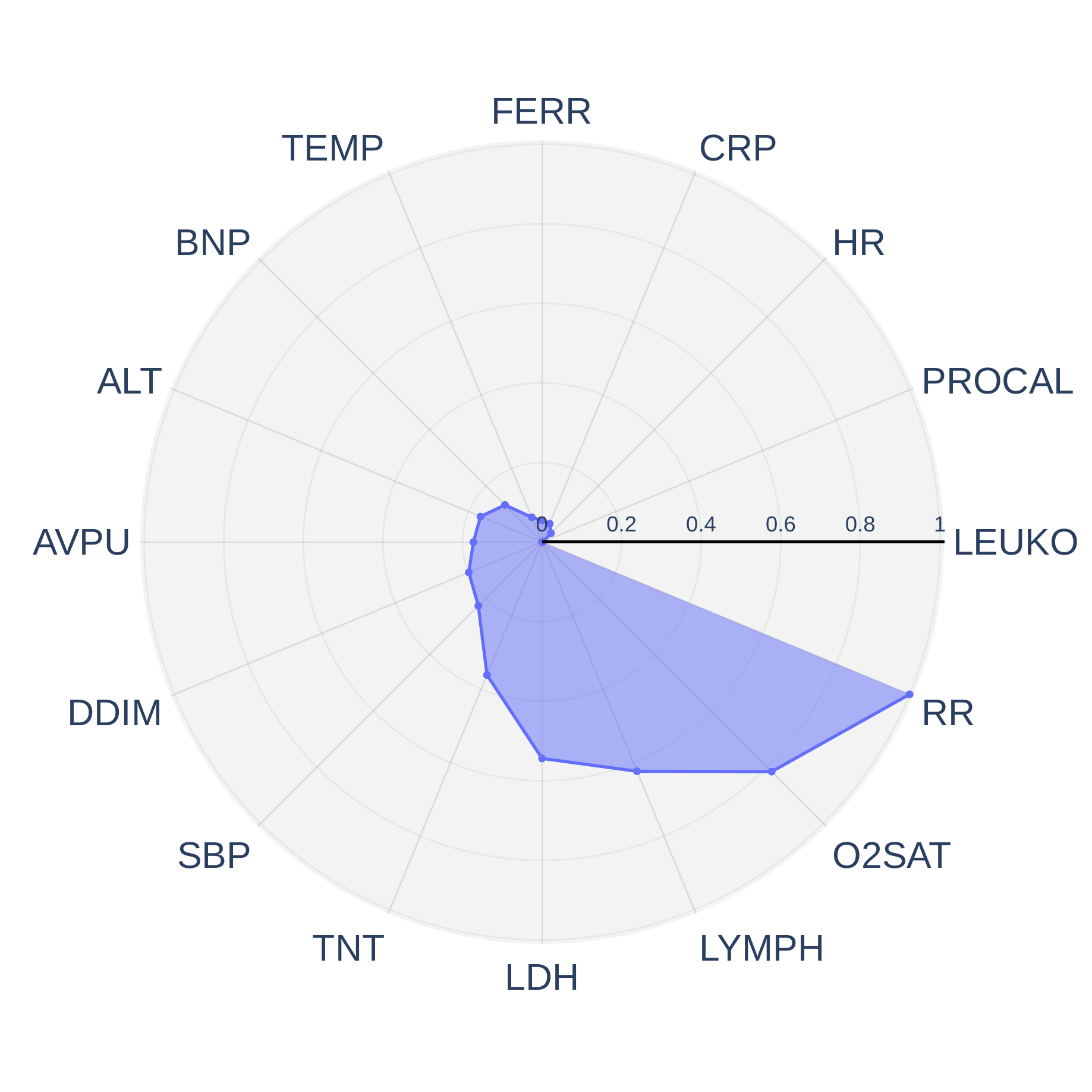}
\caption{ Feature importance computed relative to the magnitude of the AUROC drop on the training set using 5-fold cross validation when excluding each feature.}
\label{fig:feat_imp}
\end{figure}

Figure \ref{fig:lasso_path} shows how the model coefficients change as we decrease the L1 regularization strength (from left to right) for the mortality event on the Stony Brook Hospital dataset. At a very high regularization strength (Model number 0 on the x-axis), the model coefficients are all zero. The rightmost of the figure on the other hand is equivalent to non-regularized logistic regression. As we decrease the L1 regularization penalty, the model is allowed to have larger coefficients (in absolute value). To avoid overfitting, we select the regualrization strength $\lambda$ using five-fold cross-validation. Figure \ref{fig:feat_imp} shows the most important features on the same Stony Brook Hospital dataset and mortality event. Feature importance was calculated by measuring the magnitude of the average drop in training set AUC  after removing each feature separately from the dataset using 5-fold cross validation. Following this, the values were rescaled between 0 and 1 by normalizing by the magnitude of the maximum AUC drop. From the figure, we can see that O2 saturation was the by far the most important feature. This is expected, as drops in O2 saturation level is a common sign of deterioration among COVID-19 patients.

\begin{table}
\Huge
\def\arraystretch{1.5}
\resizebox{\columnwidth}{!}{
\begin{tabular}{c @{\extracolsep{\fill}}| cccccccc}
\specialrule{.15em}{.1em}{.1em} 
 \multirow{2}{2cm}{GCS} & Ranges:  & $10.5<$ & $10.50-13.50$ & $13.50-14.50$ & $>= 14.5$\\
 \hhline{~~----} 
 & Coefficients:  &  \cellcolor{red!63} $1.58$ &  \cellcolor{red!10} $0.27$ &  \cellcolor{green!85} $-0.30$ &  \cellcolor{green!0} $0.00$
 \\  \specialrule{.15em}{.1em}{.1em} 
 \multirow{2}{2cm}{HR} & Ranges:  & $59.5<$ & $59.50-89.50$ & $89.50-93.50$ & $93.50-103.50$ & $103.50-110.50$ & $110.50-120.50$ & $>= 120.5$\\
 \hhline{~~-------} 
 & Coefficients:  &  \cellcolor{green!0} $0.00$ &  \cellcolor{green!62} $-0.22$ &  \cellcolor{green!6} $-0.02$ &  \cellcolor{red!0} $0.01$ &  \cellcolor{red!9} $0.23$ &  \cellcolor{red!21} $0.54$ &  \cellcolor{red!44} $1.11$
 \\  \specialrule{.15em}{.1em}{.1em} 
 \multirow{2}{2cm}{SBP} & Ranges:  & $78.5<$ & $78.50-89.50$ & $89.50-97.50$ & $97.50-108.50$ & $108.50-119.50$ & $119.50-154.50$ & $>= 154.5$\\
 \hhline{~~-------} 
 & Coefficients:  &  \cellcolor{red!100} $2.50$ &  \cellcolor{red!40} $1.02$ &  \cellcolor{red!17} $0.44$ &  \cellcolor{red!2} $0.07$ &  \cellcolor{green!0} $0.00$ &  \cellcolor{green!100} $-0.35$ &  \cellcolor{green!7} $-0.03$
 \\  \specialrule{.15em}{.1em}{.1em} 
 \multirow{2}{2cm}{RR} & Ranges:  & $12.5<$ & $12.50-19.50$ & $19.50-21.50$ & $21.50-23.50$ & $23.50-25.50$ & $25.50-29.50$ & $>= 29.5$\\
 \hhline{~~-------} 
 & Coefficients:  &  \cellcolor{green!34} $-0.12$ &  \cellcolor{green!79} $-0.28$ &  \cellcolor{green!0} $0.00$ &  \cellcolor{red!0} $0.01$ &  \cellcolor{red!8} $0.22$ &  \cellcolor{red!19} $0.48$ &  \cellcolor{red!40} $1.01$
 \\  \specialrule{.15em}{.1em}{.1em} 
 \multirow{2}{3cm}{TEMP} & Ranges:  & $35.51<$ & $35.52-35.93$ & $35.93-36.42$ & $36.42-37.21$ & $37.21-37.44$ & $37.44-37.92$ & $>= 37.92$\\
 \hhline{~~-------} 
 & Coefficients:  &  \cellcolor{red!58} $1.46$ &  \cellcolor{red!14} $0.37$ &  \cellcolor{green!0} $0.00$ &  \cellcolor{green!30} $-0.11$ &  \cellcolor{green!4} $-0.02$ &  \cellcolor{green!0} $0.00$ &  \cellcolor{red!20} $0.52$
 \\  \specialrule{.15em}{.1em}{.1em} 
\end{tabular}}
\caption{Coefficient values for the MIMIC III mortality EventScore[All features] model.}
\label{tab:model_coef}
\end{table}

\section{Discussion}

A few systems attempted to produce interpretable machine learning models using various machine learning algorithms. For instance, RETAIN \cite{retain} uses an attention mechanism to visualize the contribution of each feature to the final score. However, one major problem with such a deep learning architecture is that the whole system cannot be verified. Each patient will produce different attention weights. The complexity of the architecture doesn't allow health care professionals to understand how the model would behave in rare cases. Moreover, such complexity prevents medical care professionals from gaining insights about the model decisions making the model a black box system. On the other hand, other systems used decision trees. In this category, for example, is a recent work that was done for the prediction of mortality among COVID-19 patients \cite{covid_trees}. Decision tree models quickly become complex to interpret, especially as the number of features increases, which increases the depth of the tree. For these reasons, linear models are still preferred in many applications since they are more interpretable. One of such systems is TREWScore \cite{trewscore}, which is an early warning score for the prediction of sepsis. Even though TREWScore and other systems use a linear model, the lack of discretization makes interpreting model coefficients more challenging. To illustrate this, let's consider a patient's decrease in temperature from 37 degrees Celsius to 36 degrees Celsius. One needs to normalize the feature and then multiply by the linear model coefficients before understanding the effect of such an increase on the total risk score. On the other hand, using EventScore, and looking at Table \ref{tab:model_coef} it can be easily identified by replacing the value of the old range with the value in the new range. Moreover, a linear model assumes that a decrease from 37 to 36 increases/decreases the risk score by the same amount as when the temperature decreases from 36 to 35, which is not reasonable, whereas, in EventScore, that's not the case. Another disadvantage of not discretizing the features is that the model will assume a unidirectional linear relationship between the feature and the clinical event. However, this is not always true, as apparent from the MEWS score. For some features, a very high or a very low value can both be signs of deterioration. 

\begin{figure}[!htp]
\centering
\includegraphics[width=1.0\linewidth]{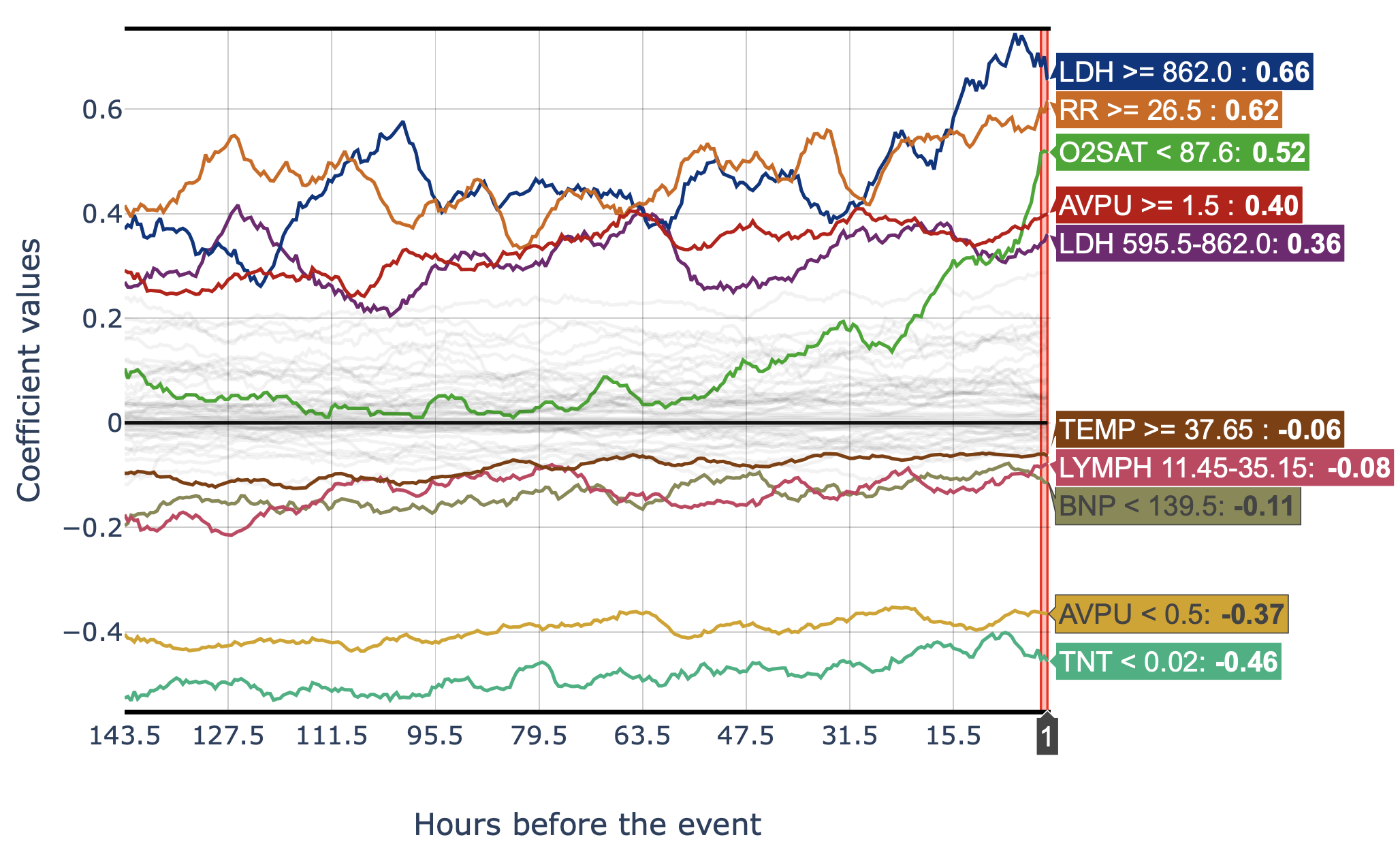}
\caption{ The average contribution of model coefficients to the final risk score of Stony Brook Hospital's positive patients as we approach their mortality event. Plots were smoothed using a moving average window of 12 hours. Patients who didn't have values at a given timestamp where excluded while computing the corresponding average at that timestamp.}
\label{fig:feat_prog}
\end{figure}

Moreover, and as shown in Figure \ref{fig:feat_prog}, discretization allows us to understand which features are raising the risk scores for the positive patients before the clinical event of interest. As we can see from the figure, one can notice a correlation between having very high LDH values and very low O2 saturation levels 32 hours before the event. Understanding such behavior between the different ranges of different features is not possible when using a baseline linear model.

Using lasso regularization gives us more insight into the development of the model. As shown in Figure \ref{fig:lasso_path}, the lasso path shows the evolution of the coefficients of the model as we decrease model complexity. This can give further insights about the model and help understand what the model will look like had we enforced stronger regularization and a simpler solution.

Another distinction between EventScore and other ICU scoring systems such as APACHE (I, II, III and IV), SAPS (I, II, III), LODS, OASIS and AutoScore is that EventScore, like MEWS and qSOFA, runs in real-time. ICU scoring systems make a single prediction after 24 hours of evaluation. While such systems are useful, they are not designed to monitor patients during their hospital stay continuously. This distinction is also reflected at training time. ICU scoring systems are trained by taking a single vector representing the patient state after 24 hours. On the other hand, EventScore is trained by taking all patient state vectors at all times. During training, those vectors are then weighted in increasing importance as we approach the clinical event of interest. 

Another advantage of EventScore is that it is a generalized algorithm that can produce an interpretable early warning system on the fly for various clinical events. While most rule-based ICU and early warning systems use a fixed list of features, EventScore can be reevaluated on a new dataset with different features and clinical events. This is especially helpful in keeping the model updated with the changing practice, populations, and clinical events. It might be more optimal to produce a specialized system for a specific hospital or a special clinical event (such as in the case of COVID-19). This is evident, for instance, from Table \ref{tab:results_final} where EventScore outperforms MEWS and qSOFA in mortality prediction on the MIMIC III dataset and the Stony Brook hospital dataset when using the same subset of features.

\section{Conclusion}

In this work, we presented EventScore, an automated early warning system for the prediction of various clinical events. We demonstrated the advantages of using EventScore over traditional early warning systems currently used in hospitals. We also demonstrated the importance of discretization by comparing EventScore to baseline logistic regression without feature discretization. 

Although EventScore is a promising scoring system, some limitations still need to be addressed. In this work, we selected some parameters, such as the decay rate of the weights of the labels of the positive patients, without further tuning. The reasoning for this is to simplify the experimental setup. However, such parameters can be further tuned using cross-validation. On the other hand, as we can see from Table \ref{tab:model_coef}, the resulting model coefficients, although interpretable, sometimes show behavior that cannot be explained. For example, GCS scores of 14 decrease the risk score while GCS values of 15 don't. This could be due to some underlying bias in the data. Although it could be considered an added advantage to EventScore,  it remains an open question whether a better performing model that adjusts to the biases in the data is better than a less performing but more interpretable model.

Another limitation of the presented methods are cases where most of the clinical events happen very early on. Without enough records, this may lead the model to learn incorrect patterns. As we saw in Table \ref{tab:results_final}, most models performed worse than random in such a scenario on the vasopressors dataset. EventScore was the only model that performed better than random when more features were added. However, such an improvement is not enough for the system to be useful for predicting the need for vasopressors. Although this impacted all the models, it indicates that early warning systems are susceptible to learning incorrect signals if the data was not filtered.

\bibliographystyle{IEEEtran}

\fontsize{9.5pt}{10.5} \selectfont \bibliography{references}

\end{document}